\documentclass[runningheads]{llncs}
\usepackage[T1]{fontenc}
\usepackage{float}
\usepackage{multirow}
\usepackage{amsfonts}
\usepackage{amsmath}
\usepackage{amssymb}
\usepackage{booktabs}
\usepackage{subcaption}
\usepackage{svg}
\usepackage{graphicx}
\usepackage{xcolor}

\begin{document}
\title{A Second-Order Attention Mechanism For Prostate Cancer Segmentation and Detection in Bi-Parametric MRI}
\titlerunning{Abbreviated paper title}

\author{Mateo Ortiz\inst{1}\orcidID{0009-0006-8343-8084} \and
Juan A. Olmos\inst{1,2}\orcidID{0000-0002-6017-0867} \and
Fabio Mart\'inez\inst{1}\orcidID{0000-0001-7353-049X}}

\authorrunning{M. Ortiz et al.}
%
\institute{
Biomedical Imaging, Vision and Learning Laboratory (BIVL$^2$ab), Universidad Industrial de Santander (UIS), 680002 Bucaramanga, Colombia \\
\email{mateo2201778@correo.uis.edu.co, jaolmosr@correo.uis.edu.co, famarcar@saber.uis.edu.co}
\and
U2IS, ENSTA, Institut Polytechnique de Paris, 91120 Palaiseau, France.
}


\titlerunning{A Second-Order attention mechanism for prostate cancer detection in MRI}

\maketitle              

\begin{abstract}
The detection of clinically significant prostate cancer lesions (csPCa) from biparametric magnetic resonance imaging (bp-MRI) has emerged as a noninvasive imaging technique for improving accurate diagnosis. Nevertheless, the analysis of such images remains highly dependent on the subjective expert interpretation. Deep learning approaches have been proposed for csPCa lesions detection and segmentation, but they remain limited due to their reliance on extensively annotated datasets. Moreover, the high lesion variability across prostate zones poses additional challenges, even for expert radiologists. This work introduces a second-order geometric attention (SOGA) mechanism that guides a dedicated segmentation network, through skip connections, to detect csPCa lesions. The proposed attention is modeled on the Riemannian manifold, learning from symmetric positive definitive (SPD) representations. The proposed mechanism was integrated into standard U-Net and nnU-Net backbones, and was validated on the publicly available PI-CAI dataset, achieving an Average Precision (AP) of 0.37 and an Area Under the ROC Curve (AUC-ROC) of 0.83, outperforming baseline networks and attention-based methods. Furthermore, the approach was evaluated on the Prostate158 dataset as an independent test cohort, achieving an AP of 0.37 and an AUC-ROC of 0.75, confirming robust generalization and suggesting discriminative learned representations.

\end{abstract}

\section{Introduction} 
Prostate cancer (PCa) is the second most common cancer and the fifth leading cause of cancer-related death among men. In 2022, over 1.5 million new cases and over 390,000 deaths were reported globally~\cite{bray2024global}. PCa detection commonly relies on the prostate-specific antigen (PSA) test, but reporting a remarked low specificity with around 36\%, leading to many false positives and unnecessary procedures~\cite{naji2018digital}. Digital rectal examination (DRE) is also used but is invasive, subjective, and limited in regions beyond rectal wall~\cite{naji2018digital}. 

Recently, multiparametric magnetic resonance imaging (mp-MRI) has demonstrated significant support in pre-biopsy assessment by localizaing  clinically significant prostate cancer (csPCa) lesions,  highlighting both vascular and morphological tissue properties~\cite{thompson2013role}. However, mp-MRI involves long acquisition times and the use of contrast agents. Alternatively, bi-parametric MRI (bp-MRI) excludes dynamic contrast enhancement (DCE) and offers comparable diagnostic accuracy, providing a practical solution for large-scale screening and routine clinical use~\cite{gatti2019prostate}. Nonetheless, in both cases mp-MRI and bp-MRI,  the interpretation and characterization of lesion findings remain subjective and expert-dependent, often leading to confusion between malignant lesions and prostatic hyperplasia~\cite{shoag2016clinical,gatti2019prostate,thompson2013role}.

Recently, U-Net based architectures have been mainly proposed for the detection of csPCa in bp-MRI due to their ability to capture multiscale contextual features ~\cite{ronneberger2015u}. Other works have integrated attention modules refining channel- and spatial-level representations during the fusion of encoder and decoder features but depending on large-scale annotated datasets ~\cite{saha2021end,hafiz2021attention}.
When trained on limited data or single-center cohorts, a common scenario, attention-based models often show reduced performance and generalization, especially on external datasets~\cite{duran2022prostattention}.

This work proposes a second-order geometric attention (SOGA) module to capture more relevant bp-MRI deep representations for the detection of csPCa lesions. The \textit{SOGA} mechanism compresses feature banks into compact second-order descriptors that summarize correlations between features. These resulting descriptors are symmetric definite positive (SPD) matrices that coexist in a smooth Riemannian manifold. Consequently, we considered Riemannian deep learning layers to learn SPD geometric descriptors. After this, the resultant SPD representations are projected into a Euclidean space and used for refining feature banks, enabling the network to learn more discriminative patterns from second-order information. The proposed \textit{SOGA} was integrated in a standard U-Net architecture and also embedded within the nnU-Net framework. The inclusion of \textit{SOGA} demonstrated to learn more discriminative and generalizable features, improving detection performance, especially in external cohort evaluation.

\section{Related Works}

Convolutional neural networks (CNNs) have been the standard tool to build computer-aided diagnosis (CAD) systems, dedicated on detecting malignant regions in medical imaging~\cite{litjens2017survey}. For csPCa detection, encoder-decoder architectures such as U-Net have become the standard, given their ability to capture malignancy patterns at the pixel level~\cite{ronneberger2015u}. Nonetheless, csPCa detection remains challenging due to the small size of many lesions, the low contrast in bp-MRI scans, and the presence of anatomically complex regions like the central zone, where benign structures may closely resemble malignant tissue~\cite{shoag2016clinical}. To overcome these limitations, attention mechanisms have been introduced to enhance feature representation. For example, \textit{Wei et al.} proposed an Attention U-Net that refines encoder-to-decoder feature maps and incorporates prostate zone information to anatomically guide the network, thereby improving detection sensitivity~\cite{wei2025enhancing}. \textit{Yang et al.} applied channel-wise attention using Squeeze-and-Excitation (SE) blocks to enhance the refinement of deep features. However, their approach focused on subregions rather than entire bp-MRI studies, limiting the network’s ability to capture broader contextual anatomical information~\cite{yang2024deep}. \textit{Li et al.} recently introduced a U-Net that also integrated anatomical information, leveraging it through a zone-aware loss function to improve lesion segmentation~\cite{li2025adaptive}. \textit{Duran et al.} proposed a dual-decoder network that jointly segments the prostate gland and lesions, using multiplicative spatial attention to infuse anatomical context into lesion segmentation~\cite{duran2022prostattention}.

Despite recent advances with attention-based methods in csPCa detection, their performance report poor generalization capabilities, \textit{i.e.,} evaluating the approach over unseen external datasets. This motivates the adoption of self-configuring frameworks like nnU-Net, which autonomously adapt pre-processing, training pipelines, and hyperparameters to the target data~\cite{isensee2021nnu}. Thus, recent approaches for csPCa detection have increasingly adopted the nnU-Net framework. For instance, Debs et al. demonstrated its superiority over a standard  U-Net~\cite{debs2024evaluation}. This has inspired new methods built upon the nnU-Net framework.~\textit{Karagoz et al.}, for example, integrated probabilistic prostate zone masks as additional input channels, emphasizing the importance of anatomical context~\cite{karagoz2023anatomically}. Although the anatomical context improves lesion detection, many existing methods continue to perform poorly in external data sets~\cite{li2025adaptive}. To overcome this, transformer-based architectures like CSwin-UNet leverage self-supervised pretraining through contrastive learning and image restoration tasks, yielding more generalizable feature representations, but with high dependency on large datasets~\cite{li2025cross}.

\section{Proposed Method}

\begin{figure}[h]
    \centering
    \includegraphics[width=1\textwidth]{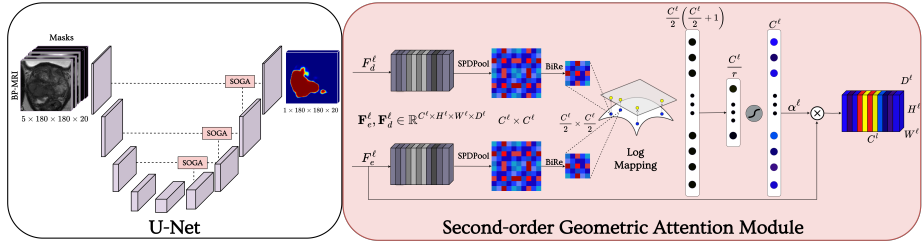}
    \caption{\textbf{Pipeline} of the proposed sencod-order geometric attention (\textit{SOGA}) module integrated in a U-Net network.}
    \label{fig:SOGA}
\end{figure}

We propose a second-order geometric attention (\textit{SOGA}) module for enhancing the encoder's skip connections by capturing channel-wise and spatial interdependencies. The proposed approach first compresses high-dimensional feature maps into compact symmetric positive definite (SPD) matrices, effectively encoding inter-channel correlations. Then, the method leverages a Riemannian processing pipeline to preserve the intrinsic geometry of the SPD manifold throughout the learning process. Finally, SPD embeddings are mapped to Euclidean space, and linear projections are applied to compute channel-wise weights that recalibrate features information to be passed to following layers in the decoder. Figure~\ref{fig:SOGA} illustrates the pipeline of the proposed \textit{SOGA} module.

\subsection{U-Net-based representation}

This work considers current end-to-end segmentation architectures for the automatic delineation of csPCa lesions on bp-MRI. Among these, the U-Net remains the most widely adopted model for biomedical image segmentation due to its ability to capture multi-scale visual context via skip connections between the encoder and decoder~\cite{ronneberger2015u}. In brief, the U-Net architecture consists of an encoder that at each level $\ell$ progressively computes blocks of features denoted by $\mathbf{F}^{\ell}_{\mathbf{e}}\in\mathbb{R}^{C_{\ell}\times H_{\ell}\times W_{\ell}\times D_{\ell}}$, where $C_{\ell}$ is the number of feature channels at level $\ell$, $H_{\ell}\times W_{\ell}$ are the spatial dimensions, and $D_{\ell}$ is the depth. These layers continue until they converge to an embedded representation (the bottleneck), from which the decoder then reconstructs the lesion delineation by progressively upsampling through decoder feature blocks $\mathbf{F}^{\ell}_{\mathbf{d}}$. At each level $\ell$, skip connections transfer encoder information to complement the decoder’s contextual representations, thereby improving segmentation performance.

Recently, skip connection has been enhanced with the incorporation of attention mechanisms, which accentuate the localization of meaningful structures, while progressively suppressing irrelevant background~\cite{oktay2018attention,woo2018cbam}. This process involves recalibrating feature activations in alignment with the target task. To this end, the input feature maps from the encoder at each level $\mathbf{F}^{\ell}_{\mathbf{e}}$ are refined through attention-based skip connections, and then incorporated in the decoder path. The contribution of this work consists in the integration of a second-order geometric mechanism in the skip connections, leveraging similarities between features to learn more relevant and discriminative representations, thereby enhancing the attention of the segmentation network.

\subsection{Geometric Riemannian learning} 

In this work, the proposed representation to summarize a bank of features is based on symmetric positive definite (SPD) matrices, capturing meaningful inter-channel relationships in a compact descriptor. To achieve this, given a feature tensor \( \mathbf{F} \in \mathbb{R}^{C \times H \times W \times D} \), we first reshape \( \mathbf{F} \) into a rectangular matrix \( \mathbf{R} \in \mathbb{R}^{C \times N} \), where \( N =  H \times W \times D \), such that each row of \( \mathbf{R} \) corresponds to a single feature information. Then, the second-order descriptor is computed as: $\mathbf{X}_0 = f_{SPDPool}(\mathbf{F}) = \frac{1}{N} \mathbf{R}\mathbf{R}^\top$, resulting in a symmetric and positive definite (SPD) matrix \( \mathbf{X}_0 \in \mathbb{R}^{C \times C} \). Here, each entry $\mathbf{X}_0 (i,j)$ encodes the correlation between the $i$-th and $j$-th features,  capturing inter-channel relationships and summarizing  relevant information from the input features.

Since SPD matrices lie on a Riemannian manifold, it is necessary to employ geometry-aware operations to preserve their structure and enable learning more meaningful representations from $\mathbf{X}_0$. Here, we utilized the \textit{SPDnet}, preserving Riemannian manifold structure while learning lower-dimensional SPD discriminative representations~\cite{huang2017riemannian}. For geometric learning, this network first projects SPD matrices into more compact SPD embeddings, following a \textit{BiMap layer}, defined as: $\mathbf{X}_k = f_{\mathrm{BiMap}}(\mathbf{X}_{k-1}) = \mathbf{W}_k \, \mathbf{X}_{k-1} \, \mathbf{W}_k^\top$, where \(\mathbf{X}_{k-1} \in \mathcal{S}_{++}^{d_{k-1}}\) is the SPD matrix from the previous layer, with dimension $d_{k-1} \times d_{k-1}$, and \(\mathbf{W}_k \in \mathbb{R}_{*}^{\,d_k \times d_{k-1}}\) is the transformation matrix that generates the new SPD matrix \(\mathbf{X}_k \in \mathcal{S}_{++}^{d_k}\). Similar to conventional networks, the dimensions of the SPD matrices are progressively reduced (\(d_k < d_{k-1}\)) via this bilinear mapping.

Resulting matrices $\mathbf{X}_k$ may have eigenvalues near zero due to optimization and numerical procedures during the learning process. So, an eigenvalue rectification (\textit{ReEig}) layer is considered \cite{huang2017riemannian}. It is defined as $
\mathbf{X}_k = f_{\mathrm{ReEig}}(\mathbf{X}_{k-1}) = \mathbf{U}_{k-1} \max(\epsilon \mathbf{I}, \mathbf{\Sigma}_{k-1}) \mathbf{U}_{k-1}^{\top}$, where \(\mathbf{U}_{k-1}\) and \(\mathbf{\Sigma}_{k-1}\) matrices correspond to the eigenvectors and eigenvalues, respectively, $\displaystyle \mathbf{X}_{k-1} = \mathbf{U}_{k-1} \mathbf{\Sigma}_{k-1} \mathbf{U}_{k-1}^{\top}$. Here, \(\epsilon \in \mathbb{R}\) denotes a non-negative rectification threshold. This operation prevents non-positive eigenvalues and preserves the SPD data structure~\cite{huang2017riemannian}. The concatenation of a \textit{BiMap} and a \textit{ReEig} layer is referred to as a \textit{BiRe} block. After a sequence of \textit{BiRe} blocks, the geometric Riemannian descriptors are mapped back to an Euclidean space using the Riemannian logarithm map (\textit{LogEig layer}): $\mathbf{X}_k = f_{\mathrm{Log}}(\mathbf{X}_{k-1}) = \mathbf{U}_{k-1}\mathrm{log}(\mathbf{\Sigma}_{k-1})\mathbf{U}_{k-1}^{\top}$. The resulting matrix \( \mathbf{X}_k \), constitutes a lower-dimensional Euclidean descriptor of semantic inter-channel relationships. The resulting $\mathbf{X}_k$ is a symmetric matrix, thereby its upper triangular part is used for further processing.

\subsection{Second Order Geometric Attention module (SOGA)}

Typically, attention mechanisms in U-Net like networks are implemented to guide encoder features ($\mathbf{F}^{\ell}_{\mathbf{e}}$), while using gating features from the decoder ($\mathbf{F}^{\ell}_{\mathbf{d}}$) for injecting semantic context and together help the network focus on regions relevant for the segmentation task. Specifically, the output of an attention mechanism is $\hat{\mathbf{F}}^{\ell} = \mathbf{F}^{\ell}_{\mathbf{e}} \odot \boldsymbol{\alpha}_{\ell}$, where \(\boldsymbol{\alpha}_{\ell}\) is a learnable vector with attention coefficients that refine the encoder feature responses. These attention coefficients are typically computed from both encoder and decoder feature tensors, \(\mathbf{F}^{\ell}_{\mathbf{e}}\) and \(\mathbf{F}^{\ell}_{\mathbf{d}}\), involving common fusion operations including concatenation, multiplication, or summation.

In this work, we propose including second-order information throughout the attention process. For this purpose, at each level $\ell$ we first compute a SPD descriptor for both the encoder and decoder features banks as $\mathbf{X}_* = f_{SPDPool}(\mathbf{F}^{\ell}_{\mathbf{*}})$, with $*\in \{\mathbf{e},\mathbf{d} \}$.  
These descriptors, of dimension $C_\ell \times C_\ell$, capture relevant information from the encoder and decoder feature banks, encoding key semantic visual patterns propagated via skip connections that are related to PCa. From the SPD descriptors \(\mathbf{X}_{\mathbf{e}}\) and \(\mathbf{X}_{\mathbf{d}}\), lower-dimensional geometric representations are obtained via two independent Riemannian SPD networks \(\varphi_{\mathbf{e}}\) and \(\varphi_{\mathbf{d}}\). Each \(\varphi_*\) consists of a \textit{BiRe} layer and a \textit{LogEig} layer which outputs a symmetric matrix of dimension $\frac{C_\ell}{2} \times \frac{C_\ell}{2}$ , and a final output layer that extracts the upper triangular part. The resulting encoder and decoder representations are concatenated to form a single geometric embedded representation:
\begin{equation} \label{eq:embedding_attention}
\mathbf{e}_\ell = \left[ \varphi_{\mathbf{e}}\!\left(f_{\mathrm{SPDPool}}(\mathbf{F}^{\ell}_{\mathbf{e}})\right),\; \varphi_{\mathbf{d}}\!\left(f_{\mathrm{SPDPool}}(\mathbf{F}^{\ell}_{\mathbf{d}})\right)\right].
\end{equation}
These fused descriptors produce a vector of dimension \(C_\ell\!\left(\frac{C_\ell}{2}+1\right)\), aggregating discriminative information from the encoder and decoder via SPD descriptors and Riemannian processing. To obtain the attention coefficients, the embedding \(\mathbf{e}_\ell\) is passed through a nonlinear mapping composed of two fully connected layers:

\begin{equation} \label{eq:att_coeficients}
\mathbf{\alpha}_\ell = \delta \left(\mathbf{W}_{\ell}^{2}\left(W_{\ell}^{1}\mathbf{e}_\ell + \mathbf{b}^1_{\ell}\right)+\mathbf{b}^2_{\ell}\right).
\end{equation}
Here, $\mathbf{W}^{1}_\ell \in \mathbb{R}^{\frac{C_\ell}{r} \times \left( C_\ell \cdot \left( \frac{C_\ell}{2} + 1 \right) \right)}$ and $\mathbf{W}^{2}_\ell \in \mathbb{R}^{C_\ell \times \frac{C_\ell}{r}}$ represent the linear transformations, and $b^1_{\ell} \in \mathbb{R}^{\frac{C_\ell}{r}}$, $b^2_{\ell} \in \mathbb{R}^{C_\ell}$ are the bias terms. The hyperparameter $r$ controls the reduction ratio, aiming to balance representational capacity and parameter efficiency. We selected $r = 4$, as described in~\cite{hu2018squeeze}. The output of this process is the vector of attention coefficients that recalibrates features as $\hat{\mathbf{F}}^{\ell} = \mathbf{F}^{\ell}_{\mathbf{e}} \odot \boldsymbol{\alpha}_{\ell}$.

\section{Dataset}

\begin{figure}[!h]
    \centering
    \includegraphics[width=1\textwidth]{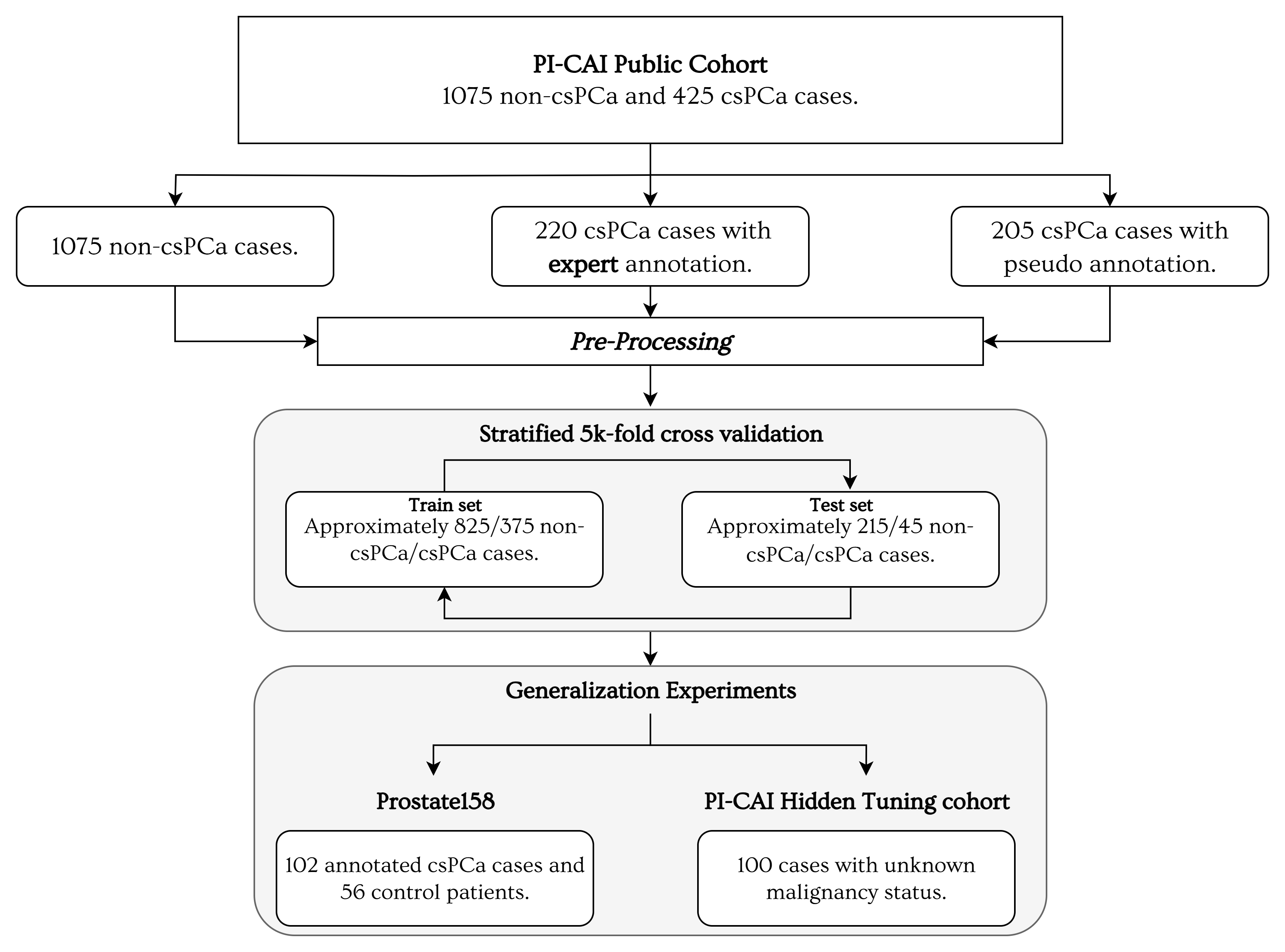}
    \caption{Datasets flowchart. The training and validation of models was performed on the public PI-CAI cohort. For generalization evaluation we considered the hidden PI-CAI Tuning cohort and the Prostate158 dataset.}
    \label{fig:datasets_flowchart}
\end{figure}

Two different datasets were included in the v0alidation of the proposed approach, among others, to assess the generalization properties of the geometrical introduced approach. Figure~\ref{fig:datasets_flowchart} describes the experimental protocol conducted to validate the approach from two considered datasets. Thereafter, we briefly describe both set of data. 

\paragraph{PI-CAI challenge dataset.}This dataset is the largest public benchmark for csPCa detection, comprising 1,500 public bp-MRI studies for public training and development.
The organizers provided delineations of the prostate's central and peripheral zones for each study. Out of the 1500 cases, 1075 are benign and 425 are biopsy-confirmed csPCa. Among the csPCa cases, 220 include experts' annotations of csPCa lesions, while the remaining 205 provide AI-generated delineations mentioned as ``pseudo-labels'', which were used exclusively for training. To address imaging variability, the dataset includes bp-MRI scans from three Dutch centers: Radboud University Medical Center (RUMC), Ziekenhuis Groep Twente (ZGT), and University Medical Center Groningen~\cite{saha2024artificial}. 

The PI-CAI dataset includes an additional hidden cohort designed for models' tuning. This cohort comprises 100 bp-MRI studies. The challenge organizers reserved the lesions' delineations for this set, and performance was assessed indirectly using metrics reported by the challenge evaluation platform.  We used this cohort to report the generalization ability of models trained on the public training cohort (see Figure~\ref{fig:datasets_flowchart}).

Pre-processing steps involved resampling the scans to a voxel resolution of $0.5\text{mm} \times 0.5\text{mm} \times 3.0\text{mm}$. Scans were then center-cropped to obtain 24 axial slices of $384 \times 384$ pixels~\cite{saha2024artificial}. A volumetric region of interest (vROI) of size $20 \times 180 \times 180$ voxels was extracted around the volumetric centroid of the prostate gland. Each input sequence, including T2-weighted images (T2W), diffusion-weighted images (DWI), and apparent diffusion coefficient maps (ADC), was independently normalized using z-score normalization.

\paragraph{PROSTATE158.} This dataset comprises bp-MRI studies from 158 patients collected at Charité University Hospital Berlin~\cite{adams2022prostate158}. Of these, 102 cases were biopsy- or surgery-confirmed csPCa, while the remaining 56 were cancer-free controls. 
The dataset is divided into 139 training and 19 testing cases, all annotated by two board-certified radiologists. Each case includes expert delineation of prostate zones (central and peripheral). The same pre-processing steps used for the PI-CAI public cohort was applied to this external cohort.
We also used this independent external cohort to evaluate the generalization of our models. The training and testing cases of PROSTATE158 were merged to form a single independent evaluation set. An overview of this workflow is provided in Figure~\ref{fig:datasets_flowchart}.

\section{Experimental Setup}
We selected a standard 3D U-Net architecture to include the proposed \textit{SOGA} attention module. The network consists of five hierarchical levels, with next channel dimensions: [32, 64, 128, 256, 512]. At each level, this architecture has a convolutional block with two 3D convolution layers (kernel size of 3$\times3\times3$), followed by batch normalization and a ReLU activation function. Downsampling was performed using 3D max pooling (see Figure~\ref{fig:SOGA}).
We refer to this model, without attention blocks, as \textit{U-Net}.  To compare the proposed Second-Order Geometric Attention (\textit{SOGA}) we considered a baseline attention module, using Global Average Pooling. This method is refer as \textit{U-Net FOA} becuse first order representation of features.  Also, the proposed approach was compared with an architecture that included SPD pooling without the subsequent geometric processing, \textit{i.e,} without the incorporation of learnable BiRE blocks. This method is refered as \textit{U-Net SOA} beacuse the second order description of features.

In the proposed \textit{SOGA} method, a rectification threshold of $\epsilon = 10^{-4}$ was applied in the ReEig layers. RMSprop with a learning rate of $10^{-4}$ was used for non-Riemannian parameters, while BiMap weights were optimized via gradient descent on the Stiefel manifold with a learning rate of $10^{-1}$ \cite{huang2017riemannian}.  All models were trained for 200 epochs using the sum of Dice Loss and Binary Cross-Entropy as the loss function.  In addition, we conducted experiments by integrating these methods into the nnU-Net framework~\cite{isensee2021nnu}. Furthermore, 
transformer-based models: UNETR~\cite{hatamizadeh2022unetr} and Swin UNETR~\cite{hatamizadeh2021swin} were adhered to the nnU-Net framework for comparison with state-of-the-art attention-based methods.  In contrast to approaches relying solely on the original U-Net architecture, these nnU-Net-based methods incorporate automated self-configuring capabilities,  such as dynamic data adaptation, offering a more robust and generalizable training pipeline~\cite{isensee2021nnu}.

\paragraph{Evaluation.} To evaluate the performance of studied methods in segmentation and detection of csPCa, multiple metrics were considered, providing patient-level and lesion-level performance. Before computing metrics, detection maps and confidence scores (maximum prediction values from these maps) were obtained using PICAI's recommended post-processing \cite{saha2024artificial}. We considered the following evaluation metrics: the AUC-ROC curve, the Average Precision (AP) with an IoU of at least 0.1, following literature recommendations~\cite{saha2024artificial}. We also included the Sensitivity at 1 False Positive per patient (Sen@1FP) score and the the Dice Similarity Coefficient (DSC) for segmentation performance.

In the PI-CAI challenge experiments, we followed the official 5-fold cross-validation using the predefined splits provided by the organizers~\cite{saha2022pi}. Ensemble predictions for the PI-CAI Hidden Tuning and Prostate158 cohorts were obtained by averaging the softmax outputs across folds.

\section{Results} 

\subsection{Detection in PI-CAI Public Cohort}

Results in the PI-CAI Public Cohort is reported in Table~\ref{tab:results_picai_public}. Interestingly, the integration of the \textit{SOA} module (\textit{U-Net SOA}) achieved the highest overall performance, while the proposed \textit{SOGA} module (\textit{U-Net SOGA}) demonstrated consistent improvements across all evaluation metrics when compared to both the baseline U-Net and its FOA-enhanced counterpart. This integration of \textit{SOGA} module increased the \textit{U-Net} performance
in AUC-ROC from 0.76 to 0.82 (+$7.9\%$), indicating a better capability of the model to distinguish between csPCa and non-csPCa cases. Besides, a wide improvement was observed analyzing the Sen@1FP score, increasing from 0.61 to 0.74 (+$21.3\%$). Additionally, the DSC increased by 0.08 (+$23.5\%$) and showed a reduction in its standard deviation.

\begin{table}[h!]
\centering
\caption{
\textbf{Performance on PI-CAI public cohort.}
Comparison between the proposed \textit{SOGA} module (marked with $^*$) and attention-based models in a U-Net configuration (top) and using the nnU-Net framework (bottom). The best models from both configurations have their performance highlighted in \textbf{bold}.
}
\label{tab:results_picai_public}
\begin{tabular}{lcccc}
\toprule
\textbf{Model}               & \textbf{AUC-ROC} & \textbf{AP}      & \textbf{Sen@1FP} & \textbf{DSC}     \\
\midrule
U-Net                        & 0.76 ± 0.03      & 0.31 ± 0.05      & 0.61 ± 0.14      & 0.34 ± 0.09      \\
U-Net FOA                    & 0.70 ± 0.04      & 0.21 ± 0.08      & 0.49 ± 0.11      & 0.25 ± 0.08      \\
U-Net SOA                   & 0.\textbf{85 ± 0.02}      & \textbf{0.38 ± 0.05}      & \textbf{0.73 ± 0.07}      & 0.22 ± 0.26      \\
U-Net SOGA$^*$              & 0.82 ± 0.03 & 0.35 ± 0.08 & 0.74 ± 0.06 & 0.42 ± 0.04 \\
\midrule
\multicolumn{5}{l}{\itshape nnU-Net framework} \\
Baseline                    & \textbf{0.83 ± 0.04} & 0.33 ± 0.09      & 0.82 ± 0.04      & 0.47 ± 0.03      \\
UNETR                       & 0.82 ± 0.03      & 0.35 ± 0.08      & 0.82 ± 0.02      & 0.49 ± 0.03      \\
Swin UNETR                  & 0.80 ± 0.03      & 0.34 ± 0.07      & \textbf{0.83 ± 0.04} & 0.50 ± 0.03      \\
FOA                         & \textbf{0.83 ± 0.03} & 0.36 ± 0.07      & 0.81 ± 0.05      & 0.51 ± 0.04      \\
SOA                   & 0.82 ± 0.04      & 0.35 ± 0.10      &  \textbf{0.83 ± 0.07   }& 0.51 ± 0.02      \\
SOGA$^*$                    & \textbf{0.83 ± 0.03} & \textbf{0.37 ± 0.05} & 0.82 ± 0.02      & \textbf{0.52 ± 0.02} \\ 
\bottomrule
\end{tabular}
\end{table}

As expected, using nnU-Net improved the performance of U-Net based methods, demonstrating the advantage of this framework for adjusting the data given its variability and configuring the training pipelines for the detection and segmentation task. Even, when \textit{SOGA} module is integrated into the nnU-Net, it demonstrated it improves the detection performance, with an AP score of $0.37\pm 0.05$ (+$5.7\%$), and in segmentation, with a DSC score of $0.52\pm 0.02$ (+$23.8\%$). Contrary, the transformer-based models UNETR and Swin UNETR achieved lower overall performance than the proposed \textit{SOGA}, and only slightly improve the performance of the baseline nnU-Net approach in AP and DSC scores.

\subsection{Generalization evaluation}

Regarding \textbf{PI-CAI Hidden Tuning cohort}, the challenge only provided  AUC-ROC and AP scores. For this experiment, we used the  nnU-Net framework as the \textit{baseline} and evaluate performance when attention modules are incorporated, as shown in Table~\ref{tab:picai_hidden_test}. Here, AUC-ROC scores indicate a similar classification performance across all methods. Nevertheless, the incorporation of the proposed \textit{SOGA} attention module led to a notable improvement in the AP score, increasing from 0.37 to 0.46 (+$24.3\%$). This difference in detection performance is higher than observed in the PI-CAI public cohort when \textit{SOGA} was incorporated in nnU-Net, indicating that the proposed \textit{SOGA} module robust on unseen cases. This substantial enhancement highlights the method’s generalization ability in unseen cases. Moreover, Swin UNETR stood out as the top-performing approach, achieving an AP of 0.54, representing an improvement of $46\%$ compared to the baseline.

\begin{table}[!h]
    \centering
    \caption{\textbf{Performance on PI-CAI hidden tuning cohort.} Comparison of the \textit{baseline} nnU-Net, nnU-Net with \textit{FOA} attention, nnU-Net with proposed \textit{SOGA} attention module, and Swin UNETR.}   
    \begin{tabular}{l c c}
        \hline
        \textbf{Model} & \textbf{AUC-ROC} & \textbf{AP} \\
        \hline
        Baseline & \textbf{0.78} & 0.37 \\
        FOA & 0.76 & 0.40 \\
        SOGA$^*$ & 0.77 & 0.46 \\
        Swin UNETR & \textbf{0.78} & \textbf{0.54} \\
        \hline
    \end{tabular}
    \label{tab:picai_hidden_test}
\end{table}

To further assess the generalization performance, we conducted additional evaluations on an independent external cohort, \textbf{Prostate158} dataset. Results are shown in Table~\ref{tab:prostate158_results}. Here, attention-based models consistently demonstrated superior generalization performance,
with the proposed \textit{SOGA} module standing out as the top performer, followed by UNETR and Swin UNETR methods. In comparison with the baseline nnU-Net, including \textit{SOGA} leads to an improvement of 0.13 points (+$21.0\%$) in AUROC, 0.22 (+$146.6\%$) in AP, 0.29 (+$126.1\%$) in Sen@1FP, and 0.19 (+$158.3\%$) in DSC.
These results underscore the ability of \textit{SOGA} module to support detection and segmentation of csPCa lesions, more remarkably in cases from external and independent cohorts. This can be explained by the ability of the proposed method to more effectively capture visual features relevant to identifying csPCa lesions, which allows for better performance in novel external, and independent datasets.

\begin{table}[H]
\centering
\caption{\textbf{Performance on the Prostate158 dataset}. Comparison of the \textit{baseline} nnU-Net and attention-based methods adhered into this framework.
}
\label{tab:prostate158_results}
\begin{tabular}{lcccc}
\toprule
\textbf{Model} & \textbf{AUC-ROC} & \textbf{AP} & \textbf{Sen@1FP} & \textbf{DSC} \\
\midrule
Baseline        & 0.62 & 0.15 & 0.23 & 0.12 \\
UNETR           & 0.73 & 0.29 & 0.38 & 0.23 \\
Swin UNETR      & 0.70 & 0.27 & 0.38 & 0.23 \\
FOA             & 0.67 & 0.18 & 0.31 & 0.18 \\
SOA             & 0.72 & 0.29 & 0.43 & 0.24 \\
SOGA$^*$        & \textbf{0.75} & \textbf{0.37} & \textbf{0.52} & \textbf{0.31} \\

\bottomrule
\end{tabular}
\end{table}

Additionally, we calculated the models' performance differentiating by lesions' size. To this end, lesions in the \textit{Prostate158} dataset were stratified by lesion volume as: \textit{small} (< 931~$mm^3$), \textit{medium} (931–2337~$mm^3$), and \textit{large} (>2337~mm$^3$), using the 33rd and 66th percentiles as thresholds. As shown in Table~\ref{tab:attention_by_size_compact}, the proposed \textit{SOGA} module consistently outperformed baseline nnU-Net and alternative attention-based methods across all lesion sizes. 
For small lesions, typically the most challenging to detect, \textit{SOGA} achieved the best performance, followed in order by SOA, UNETR, Swin UNETR, FOA, and baseline nnU-Net. In the group of medium-sized lesions, a similar order of performance is observed. Additionally, improved detection and segmentation scores are reported for all models when compared to their corresponding scores in the small lesion group. For large lesions, all methods achieved higher performance compared to medium and small lesions. While \textit{SOGA} continued to demonstrate strong overall performance, it was surpassed in AP and Sen\@1FP scores by Swin UNETR. Notably, the advantage of the proposed \textit{SOGA} was more pronounced for small and medium-sized lesions.

\begin{table}[!h]
\centering 
\caption{
\textbf{Performance across different lesion sizes.} Comparison between the \textit{baseline} nnU-Net and attention-based methods for the detection and segmentation of small, medium, and large lesions in the \textit{Prostate158} dataset.
}
\label{tab:attention_by_size_compact}
\setlength{\tabcolsep}{6pt}
\renewcommand{\arraystretch}{1.2}
\begin{tabular}{lcccccc}
\hline
\textbf{Metric} & 
\textbf{Baseline}
& \textbf{UNETR} & \textbf{S. UNETR} & \textbf{FOA} & \textbf{SOA} & \textbf{SOGA$^*$} \\
\hline
\multicolumn{7}{l}{\textit{Small} } \\
DSC     & 0.10 & 0.14 & 0.11 & 0.11 & 0.18 & \textbf{0.22} \\
AP             & 0.04 & 0.15 & 0.12 & 0.08 & 0.25 & \textbf{0.38} \\
Sen@1FP        & 0.14 & 0.28 & 0.20 & 0.17 & 0.31 & \textbf{0.50} \\
\hline
\multicolumn{7}{l}{\textit{Medium} } \\
DSC     & 0.17 & 0.28 & 0.28 & 0.24 & 0.30 & \textbf{0.33} \\
AP             & 0.26 & 0.38 & 0.36 & 0.33 & 0.42 & \textbf{0.44} \\
Sen@1FP        & 0.28 & 0.41 & 0.41 & 0.37 & 0.48 & \textbf{0.50} \\
\hline
\multicolumn{7}{l}{\textit{Large}} \\
DSC    & 0.16 & 0.26 & 0.31 & 0.24 & 0.29 & \textbf{0.40} \\
AP             & 0.24 & 0.37 & \textbf{0.47} & 0.27 & 0.35 & 0.43 \\
Sen@1FP        & 0.25 & 0.43 & 0.50 & 0.36 & 0.48 & \textbf{0.57} \\
\hline
\end{tabular}
\end{table}

\begin{figure}[H]
    \centering
        \includegraphics[width=\linewidth,height=0.8\textheight,keepaspectratio]{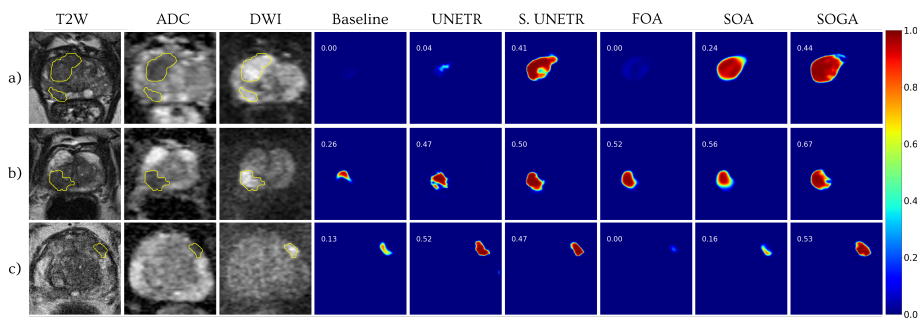}
    \caption{
        \textbf{Visualization of predictions} for three cases from the Prostate158 dataset, each presented in one row. 
        The first three columns show bp-MRI images with ground-truth annotations. 
        The predictions include DSC values.
    }
    \label{fig:qualitative_comparison}
\end{figure}

Figure~\ref{fig:qualitative_comparison} presents a visual comparison between expert annotations and models' predictions for three random cases.  In case a), there is a large lesion, delineated only by Swin UNETR, SOA and the proposed method \textit{SOGA}, with a slight advantage in segmentation by \textit{SOGA} of 3 points in the DSC, and a medium-sized lesion that was missed by all models in all predicted 3D volumes. The segmentation is not visible in the current slice, nor in the adjacent ones. Upon closer inspection, it seems the lesion appears split into two separate regions in this particular slice, which may have contributed to the models failing to detect it as a single, unified structure. Case b) presents a medium-sized lesion detected by all models. The baseline nnU-Net only captured a single portion of the lesion, in contrast with  the attention-based methods. Notably, \textit{SOGA} shows an advantage to detect lesion and also delineate better its morphology, obtaining a higher DSC. Case c) presents a small lesion in the peripheral zone. This lesion was weakly detected by the nnU-Net and SOA models, and missed entirely by the FOA model, but was successfully identified by the UNETR, Swin UNETR, and \textit{SOGA} methods. The \textit{SOGA} method provided a more precise delineation and confident detection, demonstrating the advantage of employing second-order attention over first-order statistics in proposed attention module.

\section{Discussion and conclusions remarks}

This study proposed a second-order geometric attention (\textit{SOGA}) module to capture and enhance relevant information into encoder-decoder skip connections, enhancing the segmentation and detection of PCa lesions in bp-MRI. The proposed geometric attention leverages pairwise feature relationships from Symmetric Positive Definite (SPD) descriptors within a Riemannian manifold. Extensive evaluations demonstrated the advantage of including the proposed attention in standard U-Net and nnU-Net architectures, even with respect to alternative attention-based models using transformer architectures.  This indicated that the second-order descriptors along with the geometric processing utilized in the \textit{SOGA} module effectively captured relevant patterns of PCa and improved the robustness of the baseline detection and segmentation architectures. The proposed SOGA report a remarked computational cost, during training, associated with operations on Riemannian manifolds, including eigenvalue decomposition and geometry-aware bilinear mapping.

The integration of second-order attention mechanisms over the nnU-Net led to performance generalization improvements over baseline architectures, further improving the performance on external data, obtaining an AUC-ROC of $0.83$, Sen@1FP of $0.74$, and a DSC of $0.42$. Notably, the proposed method’s advantage over the baseline and attention‑based approaches was even more pronounced on an entirely independent cohort, surpassing by 21.0\% in AUC-ROC, 126.1\% in Sen@1FP, and 158.3\% in DSC, the baseline nnU-Net model. This demonstrates its ability to capture relevant, robust features of prostate cancer and maintain a robust performance in new scenarios. This is significant for future real clinical adoption in new samples and at different clinical centers. Besides, the proposed approach offers a substantial contribution to the detection and segmentation of small lesions, key on early detection, which constitutes the most difficult challenges for both expert radiologists and deep learning models.

Future work should explore integration in novel detection frameworks, studying the relevance of geometric processing within attention mechanisms.  In line with generalization results, it is also important to evaluate the proposed module's capability in data-constrained scenarios and conduct additional studies expanding the number of datasets.


\end{document}